\documentclass{article}

\usepackage{microtype}
\usepackage{graphicx}
\usepackage{subcaption}
\usepackage{enumitem}
\usepackage{booktabs} % for professional tables

\usepackage{hyperref}

\usepackage[preprint]{icml2026}

\usepackage{amsmath}
\usepackage{amssymb}
\usepackage{mathtools}
\usepackage{amsthm}

\usepackage[capitalize,noabbrev]{cleveref}

\theoremstyle{plain}

\theoremstyle{definition}

\theoremstyle{remark}

\usepackage[textsize=tiny]{todonotes}

\icmltitlerunning{MicroFuse: Protein-to-Genome Expert Fusion for Microbial Operon Reasoning}

\begin{document}

\twocolumn[
  \icmltitle{MicroFuse: Protein-to-Genome Expert Fusion for Microbial Operon Reasoning}



  \icmlsetsymbol{equal}{*}

  \begin{icmlauthorlist}
    \icmlauthor{Seungik Cho}{yyy}
  \end{icmlauthorlist}

  \icmlaffiliation{yyy}{Department of Physics and Astronomy, Rice University, Texas, USA}

  \icmlcorrespondingauthor{Seungik Cho}{seungikcho@rice.edu}
  \icmlkeywords{microbial genomics, operon prediction, multimodal fusion, contrastive learning, foundation models}

  \vskip 0.3in
]



\printAffiliationsAndNotice{}  % no special notice (required even if empty)

\begin{abstract}
Predicting microbial operon co-membership requires integrating two 
complementary biological signals: protein-scale molecular identity 
and genome-context organization. While recent biological foundation 
models provide powerful representations of each view independently, 
naive concatenation of these modalities ignores a key biological 
property---protein identity and genomic context may agree when 
adjacent genes form a coherent functional module, or conflict when 
sequence similarity is misleading but genomic layout indicates 
independent regulation. We present \textbf{MicroFuse}, a 
protein-to-genome expert fusion framework that integrates 
structure-aware protein representations from ProstT5 with 
genome-context representations from Bacformer through a four-expert 
Mixture-of-Experts module (protein, genome-context, agreement, and 
conflict experts) with a learned soft router. Training combines 
binary cross-entropy with symmetric cross-modal InfoNCE alignment 
and disagreement-weighted supervised contrastive shaping. We further 
construct \textbf{OG-Operon100K}, a 100,000-pair scaffold-level 
benchmark from the OMG metagenomic corpus with biologically grounded 
positive and negative criteria. On OG-Operon100K, MicroFuse achieves the strongest AUROC, AUPRC,
mAP, and mAR among ProstT5-only, Bacformer-only, and Concat MLP
baselines. Ablations identify cross-modal contrastive alignment as the dominant
component, and a hard sequence-conflict subset reveals MicroFuse's
largest gains precisely in biologically ambiguous cases where protein
identity alone is misleading.
\end{abstract}

\section{Introduction}

Predicting whether two microbial genes belong to the same 
operon-like functional unit is a central challenge in microbial 
genome interpretation, with downstream applications spanning 
pathway discovery, natural product mining, metabolic engineering, 
and the functional characterization of uncultivated 
microbes~\cite{jacob1961genetic,overbeek1999use,price2005indirect,%
okuda2007door,naville2015computational}. Unlike standard protein 
function prediction, however, operon reasoning is inherently 
bimodal: it requires integrating the \textit{protein view}---where 
amino-acid sequence and structure determine molecular 
identity---with the \textit{genome-context view}---where gene order, 
orientation, intergenic distance, and local co-occurrence determine 
regulatory and functional coupling~\cite{hwang2024glm}.

Recent biological foundation models provide increasingly powerful 
representations of each view. Protein language models such as ESM 
and ProtTrans capture sequence-level evolutionary and structural 
regularities~\cite{elnaggar2021prottrans,lin2023esm2}, while 
structure-aware models such as ProstT5 additionally encode 3Di 
structural tokens from Foldseek, bridging sequence and structural 
identity in a bilingual latent 
space~\cite{heinzinger2024prostt5,vanKempen2024foldseek}. 
In parallel, genome-context models such as Bacformer 
contextualize genes using ordered bacterial genome neighborhoods, 
capturing co-regulation, gene order, and co-occurrence signals that 
are entirely absent from protein sequence 
alone~\cite{macwiatrak2025bacformer,cornman2025omg}. Yet despite 
this complementarity, the two modalities are typically fused through 
simple concatenation---an approach that treats all co-occurrence of 
protein and genomic evidence as equally informative, regardless of 
whether the two signals agree or conflict.

In this work, we introduce \textbf{MicroFuse}, a protein-to-genome 
expert fusion framework that explicitly models both modality 
agreement and disagreement for microbial operon reasoning. 
MicroFuse integrates frozen ProstT5 and Bacformer embeddings through 
a four-expert module---protein, genome-context, agreement, and 
conflict experts---weighted by a soft router conditioned on the 
joint modality representation. Cross-modal InfoNCE alignment 
and disagreement-weighted supervised contrastive shaping encourage 
the model to learn representations in which protein-to-genome 
evidence is explicitly reconciled rather than merely pooled. 
Our key contributions are:

\begin{itemize}
  \item We formulate \textbf{microbial operon co-membership 
  prediction} as a protein-to-genome foundation model fusion 
  problem, requiring joint reasoning over protein-scale molecular 
  identity and genome-context organization.

  \item We construct \textbf{OG-Operon100K}, a 100,000-pair 
  scaffold-level benchmark from the OMG metagenomic 
  corpus~\cite{cornman2025omg}, with biologically grounded 
  positive/negative criteria and scaffold-level splits to reduce 
  local-context leakage.

  \item We propose \textbf{MicroFuse}, a four-expert fusion 
  architecture trained with symmetric cross-modal InfoNCE 
  and disagreement-weighted supervised contrastive objectives, 
  explicitly representing both modality agreement and conflict.
\end{itemize}

\section{Method}

\begin{figure*}[t]
    \centering
    \includegraphics[width=0.98\textwidth]{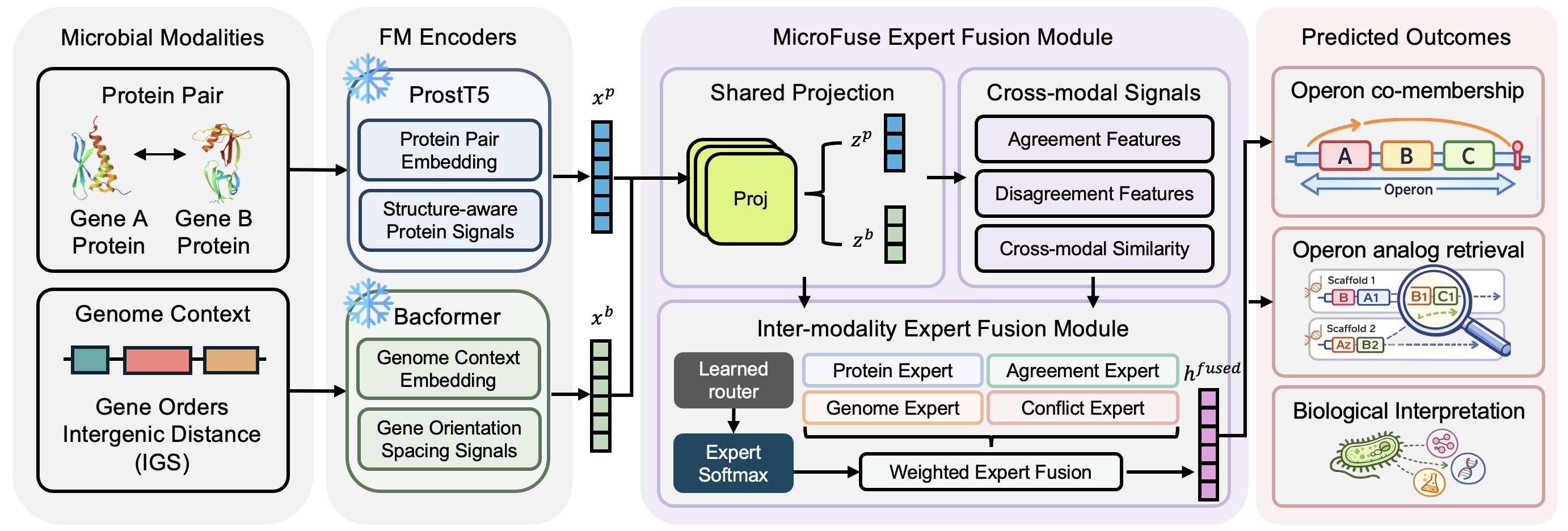}
    \caption{Overview of MicroFuse. Given a microbial gene pair, MicroFuse extracts a frozen ProstT5 protein-pair embedding and a frozen Bacformer genome-context embedding, projects both modalities into a shared latent space, and fuses them through protein, genome-context, agreement, and conflict experts. The fused representation is trained with binary task supervision, cross-modal alignment, and disagreement-weighted supervised contrastive learning for microbial operon co-membership prediction.}
    \label{fig:microfuse_overview}
\end{figure*}

Figure~\ref{fig:microfuse_overview} illustrates the MicroFuse architecture. Given
a microbial gene pair, MicroFuse takes two frozen foundation model
embeddings as input, projects them into a shared latent space,
constructs four expert representations, and fuses them to predict
operon co-membership. Training combines binary cross-entropy with
cross-modal contrastive alignment and disagreement-weighted
supervised contrastive shaping.

\subsection{Problem Formulation}

Given a pair of protein-coding genes $(g_i, g_j)$ from the same
microbial scaffold, we predict a binary label $y \in \{0,1\}$
indicating operon co-membership ($y{=}1$ for operon-like pairs,
$y{=}0$ otherwise). The model must integrate protein-level molecular
identity with genome-context organization to infer functional
coupling.

\subsection{Foundation Model Inputs}

MicroFuse uses two frozen pair-level embeddings. The
\textbf{protein embedding} $\mathbf{x}^{p}_{ij} \in \mathbb{R}^{3072}$
is derived from ProstT5~\cite{heinzinger2024prostt5}, a bilingual
protein language model over amino-acid sequences and 3Di structural
tokens from Foldseek~\cite{vanKempen2024foldseek}; two per-gene
representations and their absolute difference are concatenated per
pair. The \textbf{genome-context embedding}
$\mathbf{x}^{b}_{ij} \in \mathbb{R}^{960}$ is derived from
Bacformer~\cite{macwiatrak2025bacformer}, which contextualizes genes
using ordered bacterial genome neighborhoods to capture gene order,
orientation, and co-occurrence signals unavailable from sequence
alone~\cite{hwang2024glm}.

\subsection{Expert Fusion}

Both embeddings are projected to a shared $d{=}512$ latent space:
\begin{equation}
    \mathbf{z}^{p}_{ij} = f_p(\mathbf{x}^{p}_{ij}), \qquad
    \mathbf{z}^{b}_{ij} = f_b(\mathbf{x}^{b}_{ij}),
\end{equation}
where $f_p, f_b$ are LayerNorm--Linear--GELU--Dropout networks.
MicroFuse constructs four expert representations. Two
modality-specific experts preserve unimodal evidence:
\begin{equation}
    \mathbf{h}^{p}_{ij} = E_p(\mathbf{z}^{p}_{ij}), \qquad
    \mathbf{h}^{b}_{ij} = E_b(\mathbf{z}^{b}_{ij}).
\end{equation}
Two interaction experts explicitly model cross-modal
agreement and conflict:
\begin{align}
    \mathbf{h}^{\mathrm{agr}}_{ij} &= E_{\mathrm{agr}}\!\left(
        [\mathbf{z}^{p}_{ij} \odot \mathbf{z}^{b}_{ij};\;
         |\mathbf{z}^{p}_{ij} \odot \mathbf{z}^{b}_{ij}|]\right), \\
    \mathbf{h}^{\mathrm{conf}}_{ij} &= E_{\mathrm{conf}}\!\left(
        [|\mathbf{z}^{p}_{ij} - \mathbf{z}^{b}_{ij}|;\;
         \mathbf{z}^{p}_{ij} - \mathbf{z}^{b}_{ij}]\right).
\end{align}
A lightweight router assigns soft weights over all four experts:
\begin{equation}
    \mathbf{w}_{ij} = \mathrm{Softmax}\!\left(R\!\left(
        [\mathbf{z}^{p}_{ij};\,\mathbf{z}^{b}_{ij};\,
         |\mathbf{z}^{p}_{ij}{-}\mathbf{z}^{b}_{ij}|;\,
         \mathbf{z}^{p}_{ij}{\odot}\mathbf{z}^{b}_{ij}]\right)\right),
\end{equation}
and the fused representation and prediction are:
% 수정 — align으로 분리
\begin{align}
    \mathbf{h}_{ij} &=
        \!\sum_{k \in \{p,b,\mathrm{agr},\mathrm{conf}\}}\!\!
        w_{ij,k}\,\mathbf{h}^{k}_{ij}, \\
    \hat{y}_{ij} &= \sigma(g(\mathbf{h}_{ij})).
\end{align}

\subsection{Training Objective}

MicroFuse is trained with three objectives. \textbf{Binary
cross-entropy} supervises operon co-membership:
\begin{equation}
    \mathcal{L}_{\mathrm{BCE}} =
        -y\log\hat{y} - (1{-}y)\log(1{-}\hat{y}).
\end{equation}
\textbf{Symmetric cross-modal InfoNCE}~\cite{oord2018infonce}
aligns protein and genome-context projections. For a mini-batch
of $B$ pairs, same-pair representations are positives and all
other batch elements are negatives:
\begin{equation}
\begin{split}
    \mathcal{L}_{\mathrm{xmod}} = -\frac{1}{2B}\sum_{i=1}^{B}
    \Bigl[
        &\log\frac{e^{\mathrm{sim}(\mathbf{z}^{p}_i,\mathbf{z}^{b}_i)/\tau}}
                 {\sum_j e^{\mathrm{sim}(\mathbf{z}^{p}_i,\mathbf{z}^{b}_j)/\tau}}
      \\
      + &\log\frac{e^{\mathrm{sim}(\mathbf{z}^{b}_i,\mathbf{z}^{p}_i)/\tau}}
                 {\sum_j e^{\mathrm{sim}(\mathbf{z}^{b}_i,\mathbf{z}^{p}_j)/\tau}}
    \Bigr],
\end{split}
\end{equation}
where $\mathrm{sim}(\cdot,\cdot)$ is cosine similarity and $\tau$
is temperature. \textbf{Disagreement-weighted supervised contrastive
loss}~\cite{chen2020simclr} is applied to $\mathbf{h}_{ij}$,
encouraging same-label pairs to cluster; loss weights are the cosine
disagreement between $\mathbf{z}^{p}_{ij}$ and $\mathbf{z}^{b}_{ij}$
to up-weight biologically ambiguous examples. The final objective is:
\begin{equation}
    \mathcal{L} = \mathcal{L}_{\mathrm{BCE}}
                + \lambda_{\mathrm{xmod}}\mathcal{L}_{\mathrm{xmod}}
                + \lambda_{\mathrm{sup}}\mathcal{L}_{\mathrm{supcon}}.
\end{equation}
Only MicroFuse projection, expert, router, and classifier parameters
are trained; ProstT5 and Bacformer remain frozen throughout.

\section{Experiments}

\subsection{Experimental Setup}

\paragraph{Dataset.}
We evaluate MicroFuse on \textbf{OG-Operon100K}, constructed from
the OG/OMG metagenomic corpus~\cite{cornman2025omg}. Positive pairs
are same-strand neighboring genes separated by short intergenic
regions; negative pairs have large intergenic distances or
incompatible genomic organization; ambiguous intermediate cases are
excluded. The final dataset contains 100,000 balanced gene pairs
(50,000 positives, 50,000 negatives) with scaffold-level
train/val/test splits to reduce local-context leakage
(Table~\ref{tab:dataset}).

\begin{table}[h]
\vspace{-4pt}
\centering
\caption{OG-Operon100K dataset statistics. IGS = intergenic
spacer length (bp).}
\label{tab:dataset}
\resizebox{\columnwidth}{!}{%
\begin{tabular}{lrrrrrr}
\toprule
\textbf{Split} & \textbf{Pairs} & \textbf{Pos.} & \textbf{Neg.}
  & \textbf{Pos. rate} & \textbf{Pos. IGS} & \textbf{Neg. IGS} \\
\midrule
Train & 68,692 & 34,368 & 34,324 & 0.500 & 19.97  & 8618.85 \\
Val   & 14,169 & 6,978  & 7,191  & 0.492 & 19.67  & 8648.23 \\
Test  & 17,139 & 8,654  & 8,485  & 0.505 & 19.37  & 8769.54 \\
\bottomrule
\end{tabular}}
\vspace{-4pt}
\end{table}

\paragraph{Training details.}
All experiments are run on a single NVIDIA A100-SXM4-80GB GPU.
Embeddings are normalized using training-set statistics only.
Models are trained with AdamW~\cite{loshchilov2019adamw}, with
checkpoints selected by validation AUROC and early stopping. Full
implementation details and hyperparameters are reported in
Appendix~\ref{app:implementation}.
We report mean $\pm$ standard deviation over three random seeds
and evaluate using AUROC, AUPRC, mean average precision (mAP),
macro-F1 (mF1), macro-recall (mAR), and accuracy (ACC).

\paragraph{Baselines.}
We compare MicroFuse against three baselines:
\textit{ProstT5 only}, which trains an MLP classifier on
protein-scale embeddings alone; \textit{Bacformer only}, which
uses only genome-context embeddings; and \textit{Concat MLP},
which concatenates both embeddings and passes them through a
multilayer classifier.

\subsection{Main Results}

Table~\ref{tab:main_results} reports results on the held-out
scaffold-level test set. Single-modality models perform
substantially worse than multimodal models: ProstT5 only achieves
AUROC $0.5884 \pm 0.0060$ and Bacformer only $0.5841 \pm 0.0034$,
confirming that neither protein-scale identity nor genome-context
evidence alone is sufficient for operon co-membership prediction.

Concat MLP is a strong fusion baseline, reaching AUROC $0.6587$
and mAP $0.6598$. MicroFuse achieves the best AUROC
($\mathbf{0.6616 \pm 0.001}$), AUPRC, mAP ($\mathbf{0.6630 \pm 0.0043}$),
and mAR, while Concat MLP remains stronger on threshold-dependent
metrics (mF1, ACC). Since operon discovery is a prioritization
problem, we focus on AUROC and mAP as primary indicators of model
quality. MicroFuse improves AUROC by $+0.0028$ and mAP by $+0.0032$
relative to Concat MLP.

\begin{table}[h]
\vspace{-4pt}
\centering
\caption{Main results on OG-Operon100K. Mean $\pm$ std over
3 seeds. Best results in \textbf{bold}.}
\label{tab:main_results}
\resizebox{\columnwidth}{!}{%
\begin{tabular}{lcccccc}
\toprule
\textbf{Model} & \textbf{AUROC} & \textbf{AUPRC} & \textbf{mAP}
  & \textbf{mF1} & \textbf{mAR} & \textbf{ACC} \\
\midrule
ProstT5 only
  & $0.5884{\scriptstyle\pm.006}$
  & $0.5849{\scriptstyle\pm.005}$
  & $0.5775{\scriptstyle\pm.006}$
  & $0.5639{\scriptstyle\pm.002}$
  & $0.5644{\scriptstyle\pm.002}$
  & $0.5642{\scriptstyle\pm.002}$ \\
Bacformer only
  & $0.5841{\scriptstyle\pm.003}$
  & $0.5739{\scriptstyle\pm.004}$
  & $0.5772{\scriptstyle\pm.004}$
  & $0.5512{\scriptstyle\pm.006}$
  & $0.5571{\scriptstyle\pm.003}$
  & $0.5580{\scriptstyle\pm.003}$ \\
Concat MLP
  & $0.6587{\scriptstyle\pm.001}$
  & $0.6228{\scriptstyle\pm.004}$
  & $0.6598{\scriptstyle\pm.002}$
  & $\mathbf{0.6052{\scriptstyle\pm.005}}$
  & $0.6124{\scriptstyle\pm.002}$
  & $\mathbf{0.6137{\scriptstyle\pm.002}}$ \\
\textbf{MicroFuse}
  & $\mathbf{0.6616{\scriptstyle\pm.001}}$
  & $\mathbf{0.6319{\scriptstyle\pm.002}}$
  & $\mathbf{0.6630{\scriptstyle\pm.004}}$
  & $0.5722{\scriptstyle\pm.016}$
  & $\mathbf{0.6294{\scriptstyle\pm.002}}$
  & $0.6123{\scriptstyle\pm.002}$ \\
\bottomrule
\end{tabular}}
\vspace{-4pt}
\end{table}

\paragraph{Calibration and threshold sensitivity.}
Because MicroFuse's gains are larger on ranking metrics than on
default-threshold classification metrics (mF1, ACC), we further
analyze calibration and threshold sensitivity. MicroFuse obtains a lower Brier score than Concat MLP
($0.2380 \pm 0.0053$ vs.\ $0.3039 \pm 0.0196$). Moreover, after
threshold selection, MicroFuse reaches comparable macro-F1
($0.6126 \pm 0.0032$) and slightly higher accuracy
($0.6169 \pm 0.0054$) than Concat MLP ($0.6128 \pm 0.0039$ macro-F1,
$0.6159 \pm 0.0040$ accuracy). Thus, the lower default-threshold macro-F1 of MicroFuse appears to arise partly from threshold selection, while its ranking representation remains competitive.

\subsection{Component Ablation}

Table~\ref{tab:ablation} presents a systematic ablation over
MicroFuse components. All configurations share the same frozen
embeddings, training protocol, and evaluation metrics.

The most critical component is cross-modal contrastive alignment: its removal produces the largest single drop in the study, reducing AUROC by 0.0351 (from 0.6616 to 0.6265). The CE-only variant, which removes both contrastive objectives, yields a somewhat smaller but still substantial drop of 0.0292 AUROC (to 0.6324). Taken together, these results confirm that cross-modal contrastive alignment is the dominant driver of ranking performance, with supervised contrastive loss providing an additional but secondary contribution.

\begin{table}[h]
\vspace{-4pt}
\centering
\caption{Ablation study on OG-Operon100K. Mean $\pm$ std over
3 seeds.}
\label{tab:ablation}
\resizebox{\columnwidth}{!}{%
\begin{tabular}{lccc}
\toprule
\textbf{Configuration} & \textbf{AUROC} & \textbf{mAP} & \textbf{mF1} \\
\midrule
\textbf{MicroFuse}
  & $0.6616{\scriptstyle\pm.001}$
  & $0.6630{\scriptstyle\pm.004}$
  & $0.5722{\scriptstyle\pm.016}$ \\
w/o disagreement weighting
  & $0.6587{\scriptstyle\pm.003}$
  & $0.6587{\scriptstyle\pm.003}$
  & $0.5821{\scriptstyle\pm.009}$ \\
w/o interaction experts
  & $0.6577{\scriptstyle\pm.004}$
  & $0.6576{\scriptstyle\pm.004}$
  & $0.5786{\scriptstyle\pm.017}$ \\
w/o conflict expert
  & $0.6562{\scriptstyle\pm.003}$
  & $0.6559{\scriptstyle\pm.004}$
  & $0.5860{\scriptstyle\pm.027}$ \\
w/o supervised contrastive
  & $0.6540{\scriptstyle\pm.003}$
  & $0.6550{\scriptstyle\pm.003}$
  & $0.5929{\scriptstyle\pm.005}$ \\
CE-only
  & $0.6324{\scriptstyle\pm.019}$
  & $0.6290{\scriptstyle\pm.026}$
  & $0.5882{\scriptstyle\pm.006}$ \\
w/o cross-modal contrastive
  & $0.6265{\scriptstyle\pm.033}$
  & $0.6205{\scriptstyle\pm.037}$
  & $0.5857{\scriptstyle\pm.019}$ \\
\bottomrule
\end{tabular}}
\vspace{-4pt}
\end{table}

\subsection{Analysis}

\paragraph{Hard sequence-context conflict.}
To evaluate MicroFuse under conflicting modality evidence, we
define a hard subset using position-wise sequence identity
(truncated to 300 amino acids, aligned without gaps). Positive
pairs in the lowest quartile and negative pairs in the highest
quartile are retained (4,376 pairs, ${\approx}50\%$ positive rate).
MicroFuse achieves AUROC $0.6760 \pm 0.0133$ versus Concat MLP's
$0.6565 \pm 0.0051$ ($+0.0195$ gap), substantially larger than the
full-test margin of $+0.0028$. Single-modality models both score
below $0.572$, confirming that agreement/conflict experts provide
meaningful gains precisely when protein identity is most misleading
(Table~\ref{tab:hard}).

\begin{table}[h]
\vspace{-4pt}
\centering
\caption{Hard sequence-conflict subset results (mean $\pm$ std,
3 seeds).}
\label{tab:hard}
\resizebox{\columnwidth}{!}{%
\begin{tabular}{lccc}
\toprule
\textbf{Model} & \textbf{AUROC} & \textbf{AUPRC} & \textbf{mAP} \\
\midrule
ProstT5 only
  & $0.5666{\scriptstyle\pm.006}$
  & $0.5437{\scriptstyle\pm.003}$
  & $0.5550{\scriptstyle\pm.004}$ \\
Bacformer only
  & $0.5716{\scriptstyle\pm.011}$
  & $0.5437{\scriptstyle\pm.008}$
  & $0.5605{\scriptstyle\pm.008}$ \\
Concat MLP
  & $0.6565{\scriptstyle\pm.005}$
  & $0.6141{\scriptstyle\pm.005}$
  & $0.6556{\scriptstyle\pm.004}$ \\
\textbf{MicroFuse}
  & $\mathbf{0.6760{\scriptstyle\pm.013}}$
  & $\mathbf{0.6560{\scriptstyle\pm.023}}$
  & $\mathbf{0.6709{\scriptstyle\pm.014}}$ \\
\bottomrule
\end{tabular}}
\vspace{-4pt}
\end{table}

\paragraph{Expert usage.}
Router analysis shows that the conflict expert receives the largest
average weight ($0.393$), followed by protein ($0.252$), Bacformer
($0.231$), and agreement ($0.124$) experts. This supports the design
motivation that operon reasoning often requires resolving disagreement
between protein identity and genome context. Additional router
diagnostics are provided in Appendix~\ref{app:router_diagnostics}.

\paragraph{Cross-scaffold operon analog retrieval.}
We further evaluate whether fused representations retrieve operon-like
analogs across scaffold groups. In this retrieval setting, nearest
neighbors are selected after excluding the query scaffold group.
MicroFuse representations retrieve above-random operon-like neighbors,
suggesting that the learned space captures reusable gene-neighborhood
patterns beyond local pair classification. Detailed retrieval results are reported in Appendix~\ref{app:retrieval}.

\section{Conclusion}

We presented MicroFuse, a protein-to-genome expert fusion
framework for microbial operon reasoning that integrates
structure-aware ProstT5 protein embeddings with Bacformer
genome-context embeddings through a four-expert soft MoE module and
cross-modal contrastive alignment. We further introduced
OG-Operon100K, a 100K-pair scaffold-level benchmark from the OMG
metagenomic corpus providing a biologically grounded evaluation
for protein-to-genome foundation model fusion.

On OG-Operon100K, MicroFuse achieves the best AUROC, AUPRC, mAP, and mAR among all
baselines, with cross-modal contrastive alignment identified as
the dominant component. The largest
gains appear on the hard sequence-conflict subset, confirming that explicit agreement and conflict
modeling is most valuable when protein identity and genome context
provide competing evidence. Calibration and threshold-sweep analyses further show that
MicroFuse's weaker default-threshold macro-F1 is largely a
thresholding effect: after threshold selection, MicroFuse matches
Concat MLP in macro-F1 while slightly improving accuracy.
Together, these results suggest that protein-to-genome expert fusion
improves both ranking and classification-oriented metrics, and is
especially valuable as a representation for prioritizing microbial
operon-like functional modules.

Beyond benchmark performance, MicroFuse addresses a fundamental bottleneck in microbial biology: most environmental microbiomes remain functionally uncharacterized, and operon co-membership is the primary signal for linking genes to shared pathways. By resolving protein-genome conflicts arising from horizontal transfer and paralogy, MicroFuse can prioritize biosynthetic modules across metagenomic datasets without curated reference operons.
% =============================================================================
\bibliography{microfuse}
\bibliographystyle{icml2026}

\newpage
\appendix
\onecolumn

\section{Implementation Details}
\label{app:implementation}

All experiments were run on a single NVIDIA A100-SXM4-80GB GPU.
ProstT5 and Bacformer embeddings were precomputed and kept frozen
during downstream training. Embeddings were normalized using
training-set statistics only. Downstream models were trained with
AdamW using learning rate $8{\times}10^{-4}$ and weight decay
$10^{-4}$, with early stopping patience 14 and at most 70 epochs
with batch size 4096. MicroFuse projects both modalities to
$d{=}512$ via LayerNorm--Linear--GELU--Dropout ($p{=}0.20$)
networks. The objective uses cross-modal InfoNCE
($\tau_{\mathrm{xmod}}{=}0.12$) and supervised contrastive loss
($\tau_{\mathrm{sup}}{=}0.15$), weighted by
$\lambda_{\mathrm{xmod}}{=}0.02$ and $\lambda_{\mathrm{sup}}{=}0.03$.

\section{Calibration and Threshold Sensitivity}
\label{app:calibration}

MicroFuse achieves a lower Brier score than Concat MLP
($0.2380 \pm 0.0053$ vs.\ $0.3039 \pm 0.0196$). After threshold
selection, MicroFuse reaches macro-F1 $0.6126 \pm 0.0032$ and
accuracy $0.6169 \pm 0.0054$, comparable to Concat MLP
($0.6128 \pm 0.0039$, $0.6159 \pm 0.0040$), suggesting that the
weaker default-threshold macro-F1 is largely a thresholding effect.

\begin{figure}[h!]
    \centering
    \begin{subfigure}[t]{0.32\textwidth}
        \centering
        \includegraphics[width=\linewidth]{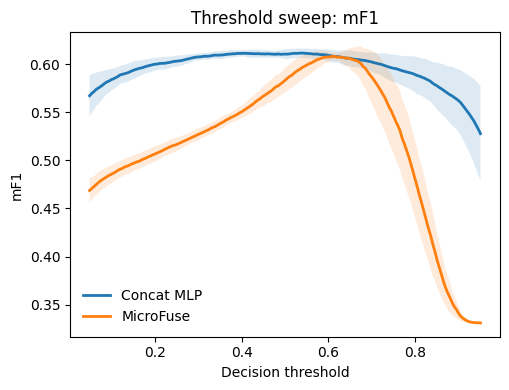}
        \caption{Macro-F1}
    \end{subfigure}
    \hfill
    \begin{subfigure}[t]{0.32\textwidth}
        \centering
        \includegraphics[width=\linewidth]{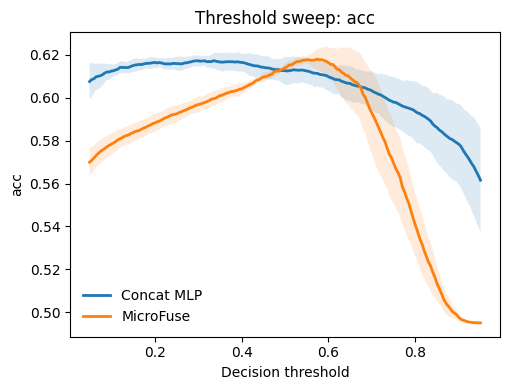}
        \caption{Accuracy}
    \end{subfigure}
    \hfill
    \begin{subfigure}[t]{0.32\textwidth}
        \centering
        \includegraphics[width=\linewidth]{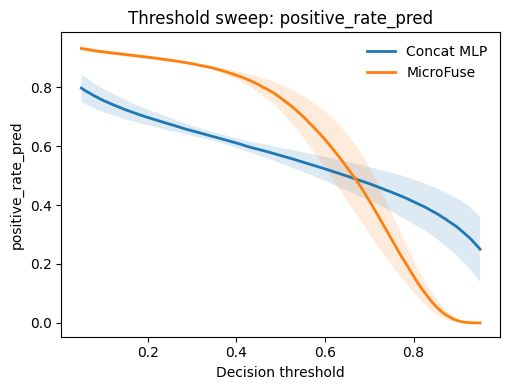}
        \caption{Predicted positive rate}
    \end{subfigure}
    \caption{Threshold sensitivity analysis. MicroFuse reaches
    comparable macro-F1 and slightly higher accuracy after
    threshold selection, suggesting that its weaker
    default-threshold macro-F1 is partly due to thresholding
    rather than representation quality.}
    \label{fig:threshold_sensitivity}
\end{figure}

\section{Router Diagnostics}
\label{app:router_diagnostics}

\begin{table}[h!]
\centering
\caption{Mean router weights on the held-out test set.}
\label{tab:expert_usage}
\begin{tabular}{lcccc}
\toprule
\textbf{Model} & \textbf{Protein} & \textbf{Bacformer}
  & \textbf{Agreement} & \textbf{Conflict} \\
\midrule
MicroFuse & 0.252 & 0.231 & 0.124 & 0.393 \\
\bottomrule
\end{tabular}
\end{table}

\section{Retrieval Analysis}
\label{app:retrieval}

\begin{table}[h!]
\centering
\caption{Cross-scaffold retrieval using fused representations.}
\label{tab:retrieval_appendix}
\begin{tabular}{ccccc}
\toprule
$k$ & \textbf{Pos.@k} & \textbf{Neg.@k}
  & \textbf{Balanced} & \textbf{Enrichment} \\
\midrule
1  & 0.573 & 0.571 & 0.572 & 1.134 \\
3  & 0.557 & 0.566 & 0.562 & 1.103 \\
5  & 0.552 & 0.565 & 0.559 & 1.093 \\
10 & 0.547 & 0.569 & 0.558 & 1.084 \\
\bottomrule
\end{tabular}
\end{table}

We evaluate retrieval of operon-like analogs across scaffold groups.
Nearest neighbors are selected after excluding the query scaffold
group. MicroFuse shows modest above-random retrieval at $k{=}1$
(positive precision $0.573$ vs.\ random ${\approx}0.494$), though
the signal diminishes at larger $k$. We treat this as qualitative
evidence of representation structure rather than a primary benchmark
result.
%%%
%%%%%%%%%%%%%%%%%%%%%%%%%%%%%%%%%%%%%%%%%%%%%%%%%%%%%%%%%%%%%%%%%%%%%%%%%%%%%%%

\end{document}